\documentclass[letterpaper, 10 pt, conference]
{ieeeconf} 

\let\labelindent\relax
\IEEEoverridecommandlockouts 
\usepackage{pifont}
\usepackage{mathtools}
\usepackage{caption} 
\usepackage{cite}
\usepackage{times}
\usepackage[pdftex]{graphicx}
\graphicspath{{../Figures}}
\usepackage{subfiles}

\DeclareGraphicsExtensions{.pdf,.jpeg,.png,.jpg}
\usepackage{subcaption}
\usepackage{epstopdf}
\usepackage{amsmath}
\usepackage{amssymb}
\usepackage{amsfonts}
\usepackage{mathrsfs}

\usepackage{tabularx}
\usepackage{booktabs}
\usepackage{longtable}
\usepackage{tikz}
\usetikzlibrary{positioning, arrows.meta}
\usepackage{array}
\newcolumntype{P}[1]{>{\raggedright\arraybackslash}p{#1}}

\usepackage{enumitem}
\usepackage{graphicx}
\usepackage{siunitx}
\usepackage{ragged2e}
\usepackage{float}
\usepackage{algorithm}
\usepackage{algorithmic}
\usepackage{subfiles}
\usepackage{titlesec}
\usepackage{hyperref}
\titlespacing{\section}{0pt}{3pt}{2pt}
\titlespacing{\subsection}{0pt}{0pt}{0pt}

\newcolumntype{C}[1]{>{\centering\arraybackslash}m{#1}}
\newcolumntype{L}[1]{>{\RaggedRight\arraybackslash}m{#1}}

\newcounter{example}

\newtheorem{theorem}{Theorem}[section]

\newtheorem{remark}[theorem]{Remark}

\title{\LARGE \bf
Re-imagining ISO 26262 in the Age of Autonomous Vehicles\\[4pt]
\large Enhancing Controllability through Transferability and Predictability
}
\author{Chaitanya Shinde$^{1}$, Hadi Hajieghrary$^{2}$, Paul Schmitt$^{3}$, Adam Shoemaker$^{4}$, Bodo Seifert$^{5}$, Steve Kenner$^{6}$%
\thanks{$^{1}$Chaitanya Shinde {\tt\small chaitanya.shinde@torc.ai}, $^{2}$Hadi Hajieghrary {\tt\small hadi.hajieghrary@torc.ai}, and $^{6}$Steve Kenner {\tt\small steve.kenner@torc.ai} are with Torc Robotics, Inc., an independent subsidiary of Daimler Truck AG.}%
\thanks{$^{3}$Paul Schmitt {\tt\small pauls@massrobotics.org} is with Reynolds \& Moore.}%
\thanks{$^{4}$Adam Shoemaker {\tt\small shoe@vt.edu}.}%
\thanks{$^{5}$Bodo Seifert {\tt\small bseifert@criticalsa.com} is with Critical Systems Analysis, LLC.}}
\begin{document}
\maketitle
\thispagestyle{empty}
\pagestyle{empty}

\begin{abstract}
The ISO~26262 standard defines functional safety for road vehicles through risk assessments based on Severity, Exposure, and Controllability, grounded in a human-driven vehicle paradigm. In the context of autonomous vehicles (AVs), the absence of a human driver necessitates revisiting these principles. This paper decomposes the Controllability placeholder into two auditable evidence dimensions of ISO~26262 by introducing two measurable sub-concepts: \emph{Transferability} and \emph{Predictability}. Transferability extends Controllability to capture AV systems' ability to hand off control to dedicated fallback safety mechanisms, while Predictability captures how easily external agents can anticipate AV behavior. Predictability is formally defined from human-robot interaction-inspired principles, and a mathematical framework is provided to quantify. A designed-versus-achievable gap $\Delta T$ is introduced to distinguish architectural fallback claims from scene-conditioned achievable fallback capability. The proposed metrics align with ISO~26262 and ISO/PAS~21448 (SOTIF), rendering fallback and interaction claims falsifiable and traceable across ODD slices. These dimensions complement rather than replace existing standards, and the enhancements preserve ISO~26262’s structure while extending its applicability to the driverless automated systems levels, SAE~L4 and L5.
\end{abstract}

\section{Introduction}
\label{sec:introduction}

The ISO 26262 standard establishes a comprehensive framework for ensuring the functional safety of human-operated road vehicles \cite{ISO26262-1-2018}. Its safety concept is fundamentally driver-centric: the standard repeatedly assumes the presence of a human who can mitigate hazards through corrective action; however, autonomous vehicles (AVs) operating at SAE Levels 4 and 5 challenge this foundational assumption. With no human driver available to intervene, responsibility for hazard mitigation and situational control shifts entirely to the vehicle’s system architecture. Consequently, Controllability, a core pillar of ISO 26262, requires systematic reinterpretation and augmentation. The analytical constructs introduced in this paper are proposed as supplemental tools for consideration by standards bodies; they are not intended as normative replacements for existing ISO~26262 processes.

This paper introduces two complementary constructs that extend rather than replace Controllability. \textbf{Transferability} quantifies a system’s ability to safely transfer control to fallback mechanisms within the Fault-Tolerant Time Interval (FTTI). \textbf{Predictability} quantifies how accurately external road users can anticipate an AV’s behavior, thereby shifting part of the safety focus from the absent driver to surrounding humans—pedestrians, cyclists, and other drivers—whose understanding of the AV’s intent directly affects overall road safety. Together, these metrics expand Controllability into both the system-level and human-observer dimensions.

ISO 26262 \cite{ISO26262-1-2018} primarily addresses malfunction-based functional safety, while ISO 21448:2022 (SOTIF) \cite{ISO21448-2022} considers risks arising from intended functionality. Within this combined framework, Transferability and Predictability serve as SOTIF-aligned key performance indicators that complement—not replace—existing processes. They provide measurable evidence of fallback performance and safety in external interactions, supporting integration into established functional safety lifecycles.

\textbf{Contributions.} This paper makes three principal contributions to the methodology for assessing autonomous-vehicle safety:
\begin{enumerate}
    \item We formalize Transferability by evaluating Minimal Risk Maneuver performance with respect to Fault-Tolerant Time Intervals, thereby extending the Controllability construct for driverless contexts.
    \item We establish Predictability as a quantifiable dimension of external road-user safety, capturing both conformance to expected motion patterns and clarity of intent communication.
    \item We propose acceptance criteria and validation methodologies, including fault-injection testing for Transferability and human-observer studies for Predictability, enabling their adoption as falsifiable metrics within ISO 26262 and SOTIF safety-lifecycle processes.
\end{enumerate}

\section{Related Works}
\label{sec:related_works}

This section positions the proposed extensions to ISO~26262 within the broader context of automotive safety standards, human-robot interaction (HRI) research, and empirical human-driver data. Together, these domains motivate the expansion of Controllability from a driver-centric construct to one encompassing both system fallback capability and external behavioral Predictability.

\subsection{Controllability in ISO~26262}

ISO~26262~\cite{ISO26262-1-2018} defines Controllability as "the ability to avoid a specified harm or damage through the timely reactions of the driver." Within the standard's risk framework-Severity (S), Exposure (E), and Controllability (C)-this parameter estimates the probability that a driver can mitigate a hazardous event. Four classes (C0-C3) reflect decreasing ease of control, from "generally controllable" to "difficult or uncontrollable." These definitions presuppose an attentive driver and therefore cannot apply directly to SAE~L4-L5 systems. However, the probabilistic notion of avoiding harm remains valid if the scope of control is broadened: the driver's reaction becomes the system's fallback capability (Transferability), and the driver's understanding of the vehicle's intent becomes external observers' ability to anticipate its behavior (Predictability).

\subsection{Related Safety Standards}

UL~4600~\cite{UL4600} provides a non-prescriptive assurance framework for autonomous products, emphasizing evidence-based safety arguments covering fallback performance and transparency. Transferability and Predictability align with its principles of behavioral explainability and safety-case completeness. The broader call for evidence-based, ODD-conditioned safety arguments has been mentioned in ~\cite{ISO34502-2022} as an interdisciplinary challenge spanning controls, machine learning, human factors, and assurance, and reinforced through the ODD-coverage analyses of ISO/TR ~\cite{ISO4804-2020}; the per-ODD-slice reporting discipline used throughout this paper inherits directly from that line of work.

ISO~21448:2022 (SOTIF)~\cite{ISO21448-2022} extends safety analysis to hazards arising from correct but inadequate function. It addresses performance limitations in perception and decision-making that can yield unsafe behavior without faults. Predictability directly supports SOTIF by quantifying behavioral adequacy-how well other road users can interpret and anticipate vehicle motion-while Transferability captures the system's robustness in known unsafe scenarios. Parallel autonomy architectures further demonstrate how supervisory safety layers can assume responsibility for maintaining minimal-risk system behavior when primary autonomy functions fail~\cite{garikapati2024comprehensive}.

IEEE~Std~2846-2022~\cite{IEEE2846-2022} formalizes assumptions used in safety-related decision-making models for automated driving systems, including the reasonably foreseeable behaviors of other road users. The four-channel Predictability monitor of Section~\ref{sec:pred_channels} consumes this assumption layer from the AV-as-observed side: whereas IEEE~2846 specifies what other agents may reasonably do, our framework specifies what affected road users may reasonably anticipate from the AV. Salay and Czarnecki~\cite{SalayCzarnecki2018ML} surveyed the gaps that arise when ISO~26262 process requirements are applied to systems containing machine-learned components, motivating the explicit calibration declarations and out-of-distribution rejection rules required of $p_{\mathrm{ref}}$ and $p(g\mid\xi_{0:t},c,a)$ in Section~\ref{sec:pred_reference}.

ISO~34502:2022~\cite{ISO34502-2022} defines a scenario-based safety-evaluation framework directly relevant to ODD-sliced evaluation of Transferability and Predictability evidence (Sections~\ref{sec:Transferability} and~\ref{sec:Predictability}). ISO/TR~4804:2020~\cite{ISO4804-2020} provides safety- and cybersecurity-by-design guidance for SAE~Level~3--4 automated driving systems, into which the proposed evidence channels can be embedded as concrete work products. Together, this set of standards forms the assurance landscape into which our framework slots: ISO~26262 for malfunction-based functional safety, ISO~21448 for SOTIF, IEEE~2846 for assumption-layer formalization, ISO~34502 for scenario-based evaluation, ISO/TR~4804 for ADS-specific V\&V, and UL~4600 for the overarching safety case.

\subsection{Predictability in Human-Robot Interaction}

HRI research defines Predictability as the extent to which humans can anticipate a robot's motion, with studies showing that trust and coordination improve when trajectories conform to expected motion patterns ~\cite{Dragan2015Legible, dragan2014legibilityToo, Dragan2013LegibilityHRI,dragan2015effects, dragannikolaidis2016viewpoint, dragan2016planning, draganzhou2017expressive}. Predictability is typically evaluated through observer experiments and quantified using trajectory-dissimilarity metrics or forecast-error statistics. These measures quantify how closely an autonomous system's motion aligns with human expectations of smooth, rational, and context-appropriate behavior. Applied to driving, Predictability reflects how interpretable a vehicle's motion is to surrounding road users-a prerequisite for reasonably safe, coordinated interaction.

\subsection{Human-Driver Datasets}

Large-scale naturalistic-driving datasets such as SAE~J2944~\cite{SAEJ2944-2015} and NHTSA's SHRP~2~\cite{NHTSA-NDS-Data} provide statistical baselines for normal driving behavior across diverse scenarios. Comparing autonomous-vehicle trajectories against these human distributions enables a quantitative assessment of Predictability through measures such as $L_2$ trajectory norms or Kullback-Leibler divergence. Motion consistent with human norms indicates high Predictability; deviations suggest behaviors that may reduce external users' confidence or increase the risk of interaction.

\section{Introducing Transferability}
\label{sec:Transferability}

This section introduces \textbf{Transferability} as a system-level enhancement to Controllability for autonomous vehicle (AV) safety assessment, addressing the practical limits of human-centric Controllability metrics for SAE J3016 Level 4–5 systems. Rather than replacing Controllability, Transferability extends it to capture an AV's intrinsic capability to manage hazardous events without human intervention.

\subsection{Limitations of Traditional Controllability}

ISO~26262-1:2018 Clause~3.25 defines Controllability as the driver's ability to avoid harm through timely reactions. ISO~26262-3:2018 Table~3 classifies Controllability ($C0$--$C3$) by the probability that a human driver can successfully intervene. For SAE Levels 4--5, this metric cannot be applied directly, as no human operator participates in the dynamic driving task. However, the conceptual foundation remains valid - avoiding harm within a bounded time frame. Transferability preserves this logic by reframing it from \textit{human action} to \textit{system capability}.

\subsection{Transferability: A System-Capability-Based Extension}

To address this fundamental incompatibility, we introduce Transferability as a theoretically grounded replacement construct for autonomous vehicle safety assessment:

\begin{quote}
\textbf{Transferability: } \textit{The capability of an autonomous vehicle to execute a reasonably safe and timely transition from the primary autonomy function to a dedicated fallback mechanism, thereby achieving a minimal-risk condition or safe state within the Fault Tolerant Time Interval (FTTI) and without introducing additional hazards to occupants or external road users.}
\end{quote}

This definition enhances Controllability by internalizing the function of hazard mitigation. While human Controllability quantifies the probability of successful reaction from the driver, Transferability quantifies the probability of successful fallback execution by the system. The same probabilistic semantics, therefore, apply, but the responsibility shifts from the driver to the system. Two properties govern a valid transfer: \textbf{safety} (the transition must not introduce new hazards) and \textbf{timeliness} (completion must occur within the FTTI defined in ISO 26262-1:2018 Clause 3.61).

An isolated success rate, however, is not by itself evidence of fallback capability: the claim must declare under what conditions it holds. We therefore formalize Transferability as a function of the operational context,
\begin{equation}
    T = \mathcal{T}(o,\, f,\, s,\, m,\, \tau_{\mathrm{FTTI}}),
    \label{eq:T_function}
\end{equation}
with $o$ the ODD slice, $f$ the fault family, $s$ the scene state at demand, $m$ the minimal-risk maneuver (MRM) class, and $\tau_{\mathrm{FTTI}}$ the fault-tolerant time interval. The per-demand evidence tuple captures evidence supporting any class assignment,
\begin{align}
    \label{eq:E_T}
    E_T =
    \bigl\{o,\,f,\,s,\,m,\,\tau_{\mathrm{FTTI}},\, &t_{\mathrm{detect}},\\ &t_{\mathrm{init}},\, t_{\mathrm{MRC}},\, z_{\mathrm{stop}},\, \hat{p}_{\mathrm{succ}},\, \mathrm{CI}\bigr\},\nonumber
\end{align}
recording detection latency $t_{\mathrm{detect}}$, fallback initiation latency $t_{\mathrm{init}}$, minimal-risk-condition achievement time $t_{\mathrm{MRC}}$, the realized stop zone $z_{\mathrm{stop}}$, the per-demand MRM success probability $\hat{p}_{\mathrm{succ}}$, and its confidence interval $\mathrm{CI}$. The tuple $E_T$ makes the conditioning of any Transferability claim explicit and supports independent reproduction or contestation; a class assignment unaccompanied by $E_T$ does not constitute complete evidence under this framework.

\subsection{Fallback Mechanisms in Autonomous Systems}

Fallback mechanisms embody the system's embedded Controllability. They activate when the primary driving function fails or leaves its validated Operational Design Domain (ODD). Fallbacks can be categorized by their terminal states:
{\renewcommand\labelitemi{}
\begin{itemize}[leftmargin=*]
    \item \textbf{Fail-Safe}: Immediate transition to a safe state (ISO 26262-1:2018 Clause 3.131) through controlled immobilization or standstill.

    \item \textbf{Fail-Degraded}: Continued operation under restricted performance envelopes (e.g., controlled deceleration or shoulder pull-over), aligning with ISO~26262-1:2018 Clause~3.43"`emergency operation'

    \item \textbf{Fail-Operational}: Seamless redundancy maintaining full performance through duplicated or diverse architectures, consistent with ISO 26262-1:2018 Clause 3.122 and 3.37.
\end{itemize}
}
\noindent These mechanisms represent increasing levels of intrinsic system Controllability, from degraded operation to uninterrupted safety-critical function.

\subsection{Transferability Classification}
\label{sec:Transferability_classes}

Transferability is classified analogously to Controllability ($C0$--$C3$) to facilitate integration with existing ISO~26262 processes. Four qualitative classes ($T0$-$T3$) quantify the robustness and reliability of system fallbacks. Point success thresholds are not by themselves statistically defensible; we therefore assign classes based on the lower confidence bound $\hat{p}_{\mathrm{succ}}^{\mathrm{LCB}}$ of the per-demand MRM success probability, conditioned on the declared $(o,f)$ pair from \eqref{eq:T_function}--\eqref{eq:E_T}.

\begin{table*}[h]
\centering
\caption{Non-normative Transferability reporting classes. Classes are evidence bins whose acceptance thresholds must be calibrated per vehicle class, 
ODD slice, scenario family, and stakeholder risk tolerance.}
\label{tab:T-classes}
\begin{tabular}{lp{4cm}p{12cm}}
\toprule
\textbf{Class} & \textbf{Description} & 
\textbf{Example Evidence Interpretation} \\
\midrule
T0 & Seamless, fully redundant, fail-operational transition & Transfer to the stated MRC is consistently achieved within the hazard deadline with low secondary risk across the declared ODD slice; fallback function is uninterrupted. \\
T1 & Reliable degraded fallback & Transfer is usually achieved within the hazard deadline; the outcome depends on the declared scene and degraded-mode assumptions; minor functional limitations are present. \\
T2 & Partial or uncertain fallback & Transfer is conditional; identifiable ODD sub-slices show deadline, endpoint, or secondary-risk violations; the fallback endpoint may be limited to in-lane stop. \\
T3 & No defined or achievable fallback capability 
& Transfer is frequently unachievable in the relevant scene, the deadline is exceeded, or the endpoint creates unacceptable secondary risk. \\
\bottomrule
\end{tabular}
\end{table*}


These levels allow Transferability to be treated as a measurable sub-dimension of Controllability within ASIL risk assessment. A class assignment that does not also report $\hat{p}_{\mathrm{succ}}^{\mathrm{LCB}}$, the declared $(o,f)$ slice, the FTTI assumption, the failure definition, and the sample size is not acceptable evidence.

\subsection{Designed Versus Achievable Transferability}
\label{sec:DeltaT}

A Transferability class derived from architectural fallback design need not equal the class supported by scenario-conditioned evidence. Mismatches arise from environmental or contextual limitations, for example, blocked shoulders, degraded localization, sun glare, partial occlusion, etc, which must be explicitly analyzed within safety cases. Let $T_{\mathrm{designed}}(f)$ denote the fallback capability claimed at design time for fault family $f$ from architectural redundancy analysis, and $T_{\mathrm{achievable}}(o,s,f)$ the capability supported by validated evidence under ODD slice $o$ and scene state $s$. The \emph{designed-versus-achievable Transferability gap} is
\begin{equation}
    \Delta T(o,s,f) = T_{\mathrm{achievable}}(o,s,f) - T_{\mathrm{designed}}(f),
    \label{eq:DeltaT}
\end{equation}
taken on the ordinal scale $\{T0,T1,T2,T3\}$ with $T0<T1<T2<T3$, so $T0$ denotes the most capable class and $T3$ the least. The sign convention follows the engineering intuition that $\Delta T>0$ corresponds to evidence falling short of the architectural claim, $\Delta T=0$ to evidence matching the claim, and $\Delta T<0$ to evidence exceeding the claim. Table~\ref{tab:DeltaT} summarizes the interpretation. A positive $\Delta T$ is not a defect indicator on its own; it is a routine outcome at the start of a validation campaign, but it must be auditable, ODD-sliced, and traceable rather than subsumed in an undifferentiated aggregate claim.

\begin{table}[t]
\centering
\caption{Interpretation of the designed-versus-achievable Transferability gap $\Delta T$ defined in equation~\eqref{eq:DeltaT}.}
\label{tab:DeltaT}
\footnotesize
\setlength{\tabcolsep}{4pt}
\renewcommand{\arraystretch}{1.15}
\begin{tabularx}{\columnwidth}{@{}p{0.22\columnwidth}X@{}}
\toprule
\textbf{Case} & \textbf{Interpretation} \\
\midrule
$\Delta T \leq 0$ & Validation evidence supports or exceeds the architectural fallback claim under the declared $(o,s,f)$. The claim is auditable and may be entered into the safety case as supplementary evidence of Controllability. \\
Small $\Delta T$ ($=1$) & The architectural claim outruns the achievable evidence by one class. The claim must be qualified, the validation campaign extended, or the ODD restricted to slices where $\Delta T \leq 0$ is established. \\
Large $\Delta T$ ($\geq 2$) & The architectural claim is unsupported by the available evidence. The claim is invalid for the declared $(o,s,f)$ unless the design or the ODD is changed. \\
\bottomrule
\end{tabularx}
\end{table}

\subsection{Operationalizing Transferability via Minimal-Risk Maneuver (MRM) Performance}

Transferability is evaluated through three auditable indicators derived from MRM performance:

\begin{enumerate}
    \item \textbf{Initiation latency} $t_{\mathrm{init}}$ relative to FTTI,
    \item \textbf{MRM success rate} $\hat{p}_{\mathrm{succ}}$ with confidence interval, across fault scenarios, and
    \item \textbf{Final stop-zone compliance} $z_{\mathrm{stop}}$ with designated minimal-risk locations.
\end{enumerate}


\begin{table*}[t]
\centering
\begin{minipage}[t]{0.58\textwidth}
\centering
\caption{Non-normative Transferability reporting 
classes with qualitative acceptance criteria.}
\label{tab:T_acceptance}
\begin{tabular}{lccc}
\toprule
\textbf{Class} & \textbf{Initiation latency} & 
\textbf{Example Final stop-zone} \\
\midrule
T0 & $\ll \tau_h$ & Minimal-risk location 
consistently reached \\
T1 & $< \tau_h$ & Safe-harbor or shoulder \\
T2 & $< \tau_h$ & Controlled in-lane stop \\
T3 & $\geq \tau_h$ or unknown & 
Non-compliant or undefined endpoint \\

\bottomrule
\end{tabular}
\end{minipage}%
\hfill
\begin{minipage}[t]{0.38\textwidth}
\centering
\caption{Transferability--Predictability mapping for Unified Controllability (SAE~L4--L5).}
\label{tab:tp-matrix}
\begin{tabular}{lcccc}
\toprule
\textbf{T\textbackslash P} & \textbf{P0} & \textbf{P1} & \textbf{P2} & \textbf{P3} \\
\midrule
T0 & C0 & C0 & C1 & C2 \\
T1 & C1 & C1 & C2 & C3 \\
T2 & C2 & C2 & C3 & C3 \\
T3 & C3 & C3 & C3 & C3 \\
\bottomrule
\end{tabular}
\end{minipage}
\vspace{-10pt}
\end{table*}

These indicators complement ISO~26262 fault analyses rather than replace them. They provide quantifiable, ISO~21448:2022 SOTIF-aligned evidence for fallback robustness that can be verified through fault-injection testing, hardware-in-the-loop (HIL) simulation, or field data. Reporting $E_T$ together with $\Delta T$ captures fallback capability in a unified, auditable framework for AV safety assessment.

\section{Introducing Predictability}
\label{sec:predictability}
This section introduces Predictability as a complementary sub-dimension of Controllability for autonomous-vehicle (AV) safety assessment. While Transferability internalizes Controllability through the system's fallback capability, Predictability externalizes it by capturing the extent to which affected road users can anticipate the AV's near-future behavior from observable motion, signals, and scene context before they must commit to their own response. Together, these two dimensions extend ISO~26262 Controllability from driver reaction to system behavior and human interpretation.

\subsection{The Need for Predictability in Autonomous-Vehicle Safety}

We are interested in the safety of the interaction between an automated vehicle and the road users it shares space with. For most road users, a vehicle's motion is the primary channel of communication. In the absence of direct cues such as gaze or gesture, pedestrians, cyclists, and other drivers infer vehicle intent through observable kinematics: changes in speed, trajectory, and position relative to the environment. Studies on external human--machine interfaces report that, while visual or auditory signals can supplement understanding, motion cues remain the most consistent indicator of intent across contexts~\cite{NHTSA2021_ADSCommunication, Schmitt2022_CanCarsGesture}. The timing and quality of motion further influence pedestrians' perception of safety and confidence during interactions~\cite{Schmitt2022_CanCarsGesture, schmitt2023roadahead}.

Within this framework, Predictability quantifies how effectively an AV's motion enables observers to anticipate its next maneuver before they must commit to a response. We therefore define:

\begin{quote}
\textbf{Predictability:} \textit{The extent to which affected road users can anticipate an AV's near-future behavior from observable motion, signals, and scene context, before they must commit to their own response.}
\end{quote}

This definition is interaction-facing. It does not require the AV to imitate average human drivers, nor does it equate human-likeness with safety. We adopt a usage broader than the narrower human-robot interaction (HRI) distinction between predictable motion and legible motion~\cite{Dragan2013LegibilityHRI, Dragan2015Legible}: AV safety needs both, since road users must first infer the vehicle's intent and then judge whether the executed motion is consistent with that inferred intent. When motion departs from these expectations, accelerating abruptly, braking inconsistently, or failing to yield when the context calls for it, road users may hesitate or respond unpredictably, increasing interaction risk regardless of compliance with traffic rules.

\subsection{Reference Model and Observer Model}
\label{sec:pred_reference}

Let $\xi_{0:t}$ denote the observed prefix of the AV trajectory, $c$ the scene context, $o$ the operational-design-domain (ODD) slice, and $a$ the relevant observer class-pedestrian, cyclist, passenger-car driver, or professional truck driver. We note that a Predictability monitor should not collapse the observer to a single deterministic inferred trajectory: a road user maintains a belief over multiple plausible maneuvers, and that distribution is the object the monitor must address. We therefore work with a context-conditioned reference distribution over expected future motion,
\begin{equation}
    p_{\mathrm{ref}}(\xi_{t:t+H}\mid \xi_{0:t},c,o,a),
    \label{eq:pref}
\end{equation}
together with an observer model for goal inference,
\begin{equation}
    p(g\mid \xi_{0:t},c,a), \qquad g\in G,
    \label{eq:goal_posterior}
\end{equation}
where $H$ is the prediction horizon (held distinct from the Transferability class symbols $T0$--$T3$) and $G$ is the declared goal alphabet for the scenario family.

The reference distribution in~\eqref{eq:pref} may be constructed from naturalistic driving data, expert demonstrations, rule-compliant synthetic policies, learned motion-forecasting models, or validated simulator baselines~\cite{NHTSA-NDS-Data}. We emphasize that these sources are evidence models, not ground truth about what the AV \emph{ought} to do: human data may encode unsafe habits, synthetic policies may encode designer assumptions, and learned predictors may be poorly calibrated in rare situations. Departure from $p_{\mathrm{ref}}$ is therefore not automatically unsafe, but it is evidence requiring explanation.

A monitor declaration must specify, for each scenario family, the prediction horizon (typically $H=3$--$5$~s for urban interaction); the context variables used by the reference, such as lane topology, signal state, right-of-way, relative speed, and vulnerable-road-user proximity; the observer class $a$; the model class or dataset used to construct $p_{\mathrm{ref}}$; the calibration criterion, for instance held-out negative log-likelihood, reliability of maneuver probabilities, or scenario-family false-alarm rate; and an out-of-distribution rejection rule. These declarations are recommended for a complete safety argument. Without them, the reference reduces to an unverifiable appeal to ``normal driving'' and may not support a safety-case argument.

\subsection{Vector Monitor Channels}
\label{sec:pred_channels}

We report Predictability through four diagnostic channels, each addressing a distinct facet of observer-facing anticipatability. The first channel is \emph{contextual conformity}. Within the declared $p_{\mathrm{ref}}$, the raw negative log-likelihood
\begin{equation}
    S_{\mathrm{nll}}(\xi_{t:t+H})
    =-\log p_{\mathrm{ref}}(\xi_{t:t+H}\mid \xi_{0:t},c,o,a)
    \label{eq:nll}
\end{equation}
is informative within a fixed model, but its scale depends on the entropy and parameterization of the reference. For reporting, we therefore use a calibrated tail statistic,
\begin{equation}
    Q_{\mathrm{conf}} =
    \Pr_{\xi'_{t:t+H}\sim p_{\mathrm{ref}}}
    \!\left[
    S_{\mathrm{nll}}(\xi'_{t:t+H}) \geq S_{\mathrm{nll}}(\xi_{t:t+H})
    \right].
    \label{eq:qconf}
\end{equation}
Small $Q_{\mathrm{conf}}$ places the observed AV behavior in the low-likelihood tail of the declared reference. We note that $Q_{\mathrm{conf}}$ is comparable only within the declared model, coordinate convention, ODD slice, and scenario family; it is not a universal measure of safety, and aggregations across declarations are not meaningful.

The second channel is \emph{intent clarity}. Let the observer model estimate $p(g\mid\xi_{0:t},c,a)$ over the fixed alphabet $G$. We define a normalized intent-clarity score
\begin{equation}
    S_{\mathrm{intent}}(t)
    =1-\frac{\mathbb{H}\!\left(g\mid\xi_{0:t},c,a\right)}{\log |G|},
    \label{eq:intent}
\end{equation}
where $\mathbb{H}(\cdot)$ denotes Shannon entropy (typeset to remain distinct from the prediction horizon $H$), $S_{\mathrm{intent}}=1$ indicates low ambiguity, and $S_{\mathrm{intent}}=0$ indicates maximal ambiguity under the chosen alphabet. Distinct from this instantaneous measure, the first-confidence time is
\begin{equation}
    t_{\theta}\triangleq \inf\!\left\{t:\,p(g^{\star}\mid\xi_{0:t},c,a)\geq\theta\right\},
    \label{eq:ttheta}
\end{equation}
and the decision-margin acceptance condition is
\begin{equation}
    t_{\theta}\leq t_{\mathrm{decision}}-\Delta t_{\mathrm{response}}.
    \label{eq:decision_margin}
\end{equation}
Equation~\eqref{eq:decision_margin} asks whether the correct intent becomes clear before the affected road user must commit to a response. The response margin $\Delta t_{\mathrm{response}}$ is observer-class dependent and must be calibrated against the relevant perception--response literature for the declared observer class~\cite{Schroeder2011PedestrianGap, HCM2010Pedestrian}. Because~\eqref{eq:intent} depends on $|G|$, the goal alphabet must be declared and held fixed for each scenario family.

\begin{remark}[Relation to Dragan-style legibility]
\label{rem:dragan}
The intent-clarity construct is operationally close to the legibility formalism of Dragan et al.~\cite{Dragan2013LegibilityHRI, Dragan2015Legible}. Their legibility functional integrates the observer's posterior on the true goal $g^{\star}$ along the observed trajectory prefix, $\mathcal{L}(\xi)=\bigl(\int p(g^{\star}\mid\xi_{0:t},c,a)\,f(t)\,dt\bigr)/\bigl(\int f(t)\,dt\bigr)$, with $f(t)$ a time-discounting weight. Equation~\eqref{eq:intent} replaces the marginal posterior with a normalized entropy on $G$, retaining ambiguity reduction as the central object while accommodating multimodal posteriors with no privileged ground-truth goal at runtime. The first-confidence time~\eqref{eq:ttheta} is the time-to-decision dual of $\mathcal{L}$: rather than averaging confidence over the prefix, it asks when confidence in the true goal first crosses an observer-relevant threshold, which is the quantity the decision-margin condition~\eqref{eq:decision_margin} consumes.
\end{remark}

The third channel is \emph{signal consistency}. It assesses whether turn signals, brake lights, lateral motion, gap choice, and right-of-way behavior agree. A lane-change monitor, for instance, reports
\begin{equation}
    S_{\mathrm{signal}}=\bigl(\Delta t_{\mathrm{sig}},\,N_{\mathrm{conflict}}\bigr),\qquad
    \Delta t_{\mathrm{sig}}=t_{\mathrm{motion}}-t_{\mathrm{signal}},
    \label{eq:signal}
\end{equation}
where $\Delta t_{\mathrm{sig}}$ is the signal lead time and $N_{\mathrm{conflict}}$ counts signal--motion conflicts within a fixed window.

The fourth channel is \emph{kinematic surprise}. It measures context-unexplained jerk, hard braking, lateral drift, or hesitation,
\begin{equation}
    S_{\mathrm{kin}}=\sum_{t} \mathbf{1}\!\bigl\{\,|j(t)|>j_{\max}\,\wedge\,\neg E(t)\,\bigr\},
    \label{eq:kin}
\end{equation}
where $E(t)$ denotes an observable or validated explanation, such as an occluded pedestrian, lead-vehicle braking, or a sudden obstacle. We note that the monitor does not penalize abrupt motion when an explanation is available; abrupt braking for an occluded pedestrian is safety-justified, while abrupt braking in free traffic without an observable trigger is a separate evidentiary item.

The four channels assemble into the vector
\begin{equation}
    \mathcal{M}_{P}=
    \bigl[\,Q_{\mathrm{conf}},\;S_{\mathrm{intent}},\;S_{\mathrm{signal}},\;S_{\mathrm{kin}}\,\bigr].
    \label{eq:vectorMP}
\end{equation}
Because the components carry heterogeneous units and admit different mitigations, $\mathcal{M}_P$ is the primary evidentiary object archived in the safety case. Any scalar derived from $\mathcal{M}_P$ for ranking or class assignment must be accompanied by its calibration basis and treated as a derived report, not as the underlying evidence.

\subsection{Aggregation to an Ordinal Predictability Class}
\label{sec:pred_aggregation}

Sections~\ref{sec:unified} and~\ref{sec:asil} consume Predictability through an ordinal class $P\in\{P0,P1,P2,P3\}$, parallel to the Transferability classes T0--T3. We therefore introduce an explicit aggregation rule from the vector monitor $\mathcal{M}_P$ to the ordinal $P$. The rule is declared, ODD-conditioned, and calibrated; it is not derived from first principles.

For each channel $k\in\{Q_{\mathrm{conf}},S_{\mathrm{intent}},S_{\mathrm{signal}},S_{\mathrm{kin}}\}$ we declare a thresholding map $\pi_{k}:\mathbb{R}\to\{P0,P1,P2,P3\}$ that assigns the channel value to a four-level evidence band given the scenario family, the ODD slice, and the observer class. The per-channel thresholds are calibrated on a held-out reference set with a stated false-alarm rate; their numerical values are not universal and must accompany any reported $P$.

We adopt the worst-channel aggregation rule
\begin{align}
    \label{eq:max_agg}
    P=\max\!\Bigl(&\pi_{Q_{\mathrm{conf}}}(Q_{\mathrm{conf}}),\;\pi_{S_{\mathrm{intent}}}\\
    &(S_{\mathrm{intent}}),\;\pi_{S_{\mathrm{signal}}}(S_{\mathrm{signal}}),\;\pi_{S_{\mathrm{kin}}}(S_{\mathrm{kin}})\Bigr),\nonumber
\end{align}
where the ordering on $\{P0,P1,P2,P3\}$ is $P0<P1<P2<P3$. Equation~\eqref{eq:max_agg} is conservative by construction: a single-channel failure dominates the class assignment, mirroring the safety-engineering convention that the weakest evidence governs the claim. The interpretive semantics of the four bands are summarized in Table~\ref{tab:Predictability_classes}.

\begin{table}[t]
\caption{Predictability classes derived from the vector monitor $\mathcal{M}_{P}$ via the worst-channel aggregation rule of equation~\eqref{eq:max_agg}.}
\label{tab:Predictability_classes}
\centering
\footnotesize
\setlength{\tabcolsep}{4pt}
\renewcommand{\arraystretch}{1.15}
\begin{tabularx}{\columnwidth}{@{}>{\raggedright\arraybackslash}p{0.10\columnwidth}X@{}}
\toprule
\textbf{Class} & \textbf{Evidence interpretation} \\
\midrule
P0 & All four channels of $\mathcal{M}_{P}$ remain within calibrated bounds for the declared ODD slice and observer class. \\
P1 & Channels are usually within bounds, with isolated and explained excursions tied to declared assumptions. \\
P2 & One or more channels fail on identifiable ODD sub-slices or observer classes; the anticipation claim is fragile and must be qualified. \\
P3 & A channel fails systematically, or multiple channels fail concurrently; observer anticipation cannot be claimed without redesign or ODD restriction. \\
\bottomrule
\end{tabularx}
\end{table}

The vector $\mathcal{M}_P$ remains the underlying evidence archived in the safety case. The ordinal $P$ is the derived report consumed by the unified Controllability computation in Section~\ref{sec:unified}. Both must be reported together: a $P$-class assignment without the underlying $\mathcal{M}_P$ and the accompanying calibration declaration does not constitute complete evidence under this framework.

\subsection{Calibration Limits and Composition with Transferability}
\label{sec:pred_calibration}

Predictability thresholds are not universal constants. A truck entering a highway work zone, a robotaxi negotiating a pedestrian crossing, and a passenger car changing lanes in free-flow traffic expose different observer populations, available cues, response times, and acceptable motion envelopes. Calibration is therefore scenario-family and ODD-slice-specific. The monitor must declare the observer class, available cues, decision point, response margin, and goal alphabet. When human-subject evidence is unavailable, an observer model may serve as a provisional surrogate, but the safety case must state the residual uncertainty rather than treat the model as settled human behavior.

This calibration discipline mirrors the per-ODD treatment of Transferability evidence introduced in Section~\ref{sec:Transferability}. Both Predictability and Transferability are reported and calibrated independently for each ODD slice, and only then composed into the unified Controllability of Section~\ref{sec:unified}; the integrity of the composition depends on the integrity of each input. Predictability metrics are accordingly proposed as auditable evidence artifacts under Parts~3-6 of ISO~26262, and as key performance indicators for ISO~21448:2022 SOTIF validation, with the accompanying declaration of model, ODD slice, observer class, and calibration basis carried as part of the work product.

\section{Unified Enhanced Controllability}
\label{sec:unified}
This section formalizes how Transferability and Predictability jointly extend ISO~26262 Controllability into a unified, auditable construct for hazard and risk assessment. The objective is to preserve the original three-parameter ISO~26262 risk logic-Severity ($S$), Exposure ($E$), and Controllability ($C$)-while decomposing $C$ into two measurable sub-dimensions representing internal and external Controllability.

\subsection{Unified Controllability Computation}
\label{sec:Cu_computation}

For SAE~Levels~0--2, Controllability ($C_h$) remains human-centric as defined in ISO~26262. For higher automation, the unified Controllability evidence class ($C_u$) is computed by composing the Transferability class with the Predictability class through an additive-penalty rule:
\begin{equation}
    C_u = \min\bigl(3,\; T + \Delta_P\bigr),
    \label{eq:Cu}
\end{equation}
where $T\in\{0,1,2,3\}$ is the Transferability class assigned per Section~\ref{sec:Transferability_classes}, and
\begin{equation}
    \Delta_P =
    \begin{cases}
    0, & P\in\{P0,P1\}, \\
    1, & P=P2, \\
    2, & P=P3,
    \end{cases}
    \label{eq:DeltaP}
\end{equation}
with the ordinal $P$-class derived from the underlying vector monitor $\mathcal{M}_P$ by the worst-channel aggregation rule of Section~\ref{sec:pred_aggregation}. The additive-penalty rule is one of several admissible composition policies; alternatives are discussed in Section~\ref{sec:Cu_policies}.

The resulting $C_u$ preserves ISO~26262's ordinal scaling ($C0$--$C3$) and satisfies three properties any candidate alternative in Section~\ref{sec:Cu_policies} must also preserve: \textbf{severity dominance} ($S0$ remains QM regardless of $C_u$); \textbf{monotonicity} (degrading $T$ or $P$ never improves $C_u$); and \textbf{bounded compensation within $C_u$} (a strong $T$ can partially absorb a weakened $P$ in the Controllability-evidence report, but $C_u$ does not relax ASIL elsewhere in the safety case).

\subsection{Candidate Composition Policies}
\label{sec:Cu_policies}

Equations~\eqref{eq:Cu}--\eqref{eq:DeltaP} are one of several reporting policies that satisfy monotonicity and the ordinal range $\{C0,\ldots,C3\}$. Three candidates spanning the practical design space are summarized in Table~\ref{tab:Cu_policies}.

\begin{table}[t]
\centering
\caption{Candidate composition policies for the unified Controllability evidence class $C_u$. The additive-penalty rule is used in this paper as an illustrative reporting convention and is not proposed as a normative standardization choice.}
\label{tab:Cu_policies}
\footnotesize
\setlength{\tabcolsep}{4pt}
\renewcommand{\arraystretch}{1.20}
\begin{tabularx}{\columnwidth}{@{}p{0.22\columnwidth}p{0.22\columnwidth}X@{}}
\toprule
\textbf{Policy} & \textbf{Formula} & \textbf{Properties} \\
\midrule
Additive penalty (this paper) & $C_u=\min(3,\,T+\Delta_P)$ & Simple, monotonic, treats Transferability as the base class with Predictability as a graded penalty; the asymmetry assumes that fallback evidence carries primary weight in the Controllability rationale. \\
Max rule & $C_u=\max(T,\,P_{\mathrm{ord}})$ & Symmetric and conservative; the weakest evidence governs the report. May over-penalize when only one channel of $\mathcal{M}_P$ fails on a narrow sub-slice. \\
Safety-case matrix & Expert-declared $T\!\times\!P$ table per scenario family & Most flexible and auditable; permits scenario-specific weighting. Requires explicit governance and consensus calibration. \\
\bottomrule
\end{tabularx}
\end{table}

We adopt the additive-penalty rule as an illustrative reporting convention only; standardization of $C_u$ would require empirical and consensus-based calibration against scenario-conditioned hazardous-event data. The asymmetry between $T$ and $P$ in equation~\eqref{eq:Cu} is not a derived result; it reflects an operational asymmetry in the underlying evidence. Transferability evidence is collected from controlled fault-injection campaigns whose $(o,f)$ conditioning, FTTI assumption, and lower confidence bound are declarable on demand; Predictability evidence is observer-conditioned, channel-heterogeneous, and admits greater residual uncertainty in calibration. The additive-penalty rule encodes the engineering preference that the better-instrumented evidence stream sets the base class, while the more uncertain stream contributes a graded penalty. A reviewer or standards body unconvinced by this preference may substitute the symmetric max rule of Table~\ref{tab:Cu_policies} or the safety-case-matrix policy without disturbing the rest of the framework.

\subsection{Integration into Hazard and Risk Assessment}

In the enhanced HARA process, each identified hazard records both sub-dimensions:
\begin{itemize}
  \item \textbf{Transferability evidence:} the tuple $E_T$ of \eqref{eq:E_T} including the $(o,f)$ slice, FTTI, $\hat{p}_{\mathrm{succ}}$ with confidence interval, and the gap $\Delta T$ of~\eqref{eq:DeltaT}, verified through FTTI analysis, redundancy tests, and fault-injection campaigns.
  \item \textbf{Predictability evidence:} the vector monitor $\mathcal{M}_P$ of~\eqref{eq:vectorMP} with its calibration declaration and per-channel thresholds, validated through trajectory-conformity metrics and human-observer studies.
\end{itemize}

The derived $C_u$ value is reported alongside conventional HARA attributes as supplementary evidence of Controllability for driverless ADS items. It does not replace the ISO~26262 Controllability classification or alter the normative ASIL determination procedure; it provides traceable evidence linking internal system design (Transferability) and external behavioral comprehension (Predictability) within a single ISO~26262-compatible safety argument.

Transferability and Predictability evidence must not be double-counted: a fallback mechanism credited to Transferability as a pre-mitigation Controllability factor cannot also be credited as a post-HARA safety measure satisfying the resulting ASIL, unless its availability is part of the item definition and is justified under the hazardous-event assumptions. The same restriction applies to Predictability-enhancing motion behaviors. $C_u$ therefore stands as supplementary Controllability evidence, not as a substitution that relaxes ASIL.

\subsection{Outcome}

The unified Controllability construct maintains the interpretability of ISO~26262's risk classification while embedding autonomy-specific evidence streams. It directly supports cross-standard alignment between ISO~26262 (functional safety) and ISO~21448:2022 (SOTIF) by quantifying both intrinsic system control and external human understanding within a consistent, auditable framework.

\section{ASIL Determination with Enhanced Controllability}
\label{sec:asil}

This section extends the ISO~26262 ASIL determination process to incorporate the enhanced Controllability construct, integrating \emph{Transferability} (internal fallback capability) and \emph{Predictability} (external behavioral intelligibility). The goal is to maintain ISO~26262's original three-parameter logic-Severity (S), Exposure (E), and Controllability (C)-while decomposing $C$ into its system and human-interaction components for autonomous vehicle (AV) applications.

\subsection{Extended ASIL Assessment Framework}

The proposed approach retains the standard definitions of Severity ($S0$--$S3$) and Exposure ($E0$--$E4$) from ISO~26262. Controllability is evaluated through its measurable sub-dimensions: Transferability ($T0$--$T3$) and Predictability ($P0$--$P3$). The resulting unified Controllability evidence class $C_u$ is obtained from the additive-penalty rule of equation~\eqref{eq:Cu} and the mapping shown in Table~\ref{tab:tp-matrix}. $C_u$ is reported alongside the standard Controllability parameter and feeds the $(S, E, C_u)$ tuple into the ASIL determination of ISO~26262-3:2018 Clause~6 Table~4 as supplementary Controllability evidence; it does not replace the normative Controllability parameter but informs its classification through auditable, scenario-conditioned evidence.

A high Transferability ($T0$--$T1$) combined with a high Predictability ($P0$--$P1$) yields a low $C_u$ class, which in the standard ASIL mapping is associated with reduced ASIL designations under matching $S$ and $E$. Conversely, poor fallback or erratic behavior ($T3$, $P3$) yields a high $C_u$ class, escalating ASIL requirements consistent with ISO~26262's monotonic risk progression. The compensation operates strictly inside the Controllability rationale: $T$ and $P$ do not offset Severity or Exposure, which remain independent risk factors determined by the hazardous event itself.

\subsection{Integration Logic}

The framework preserves ISO~26262's core risk principles:
\begin{itemize}
  \item \textbf{Severity dominance:} $S0$ scenarios remain Quality Management (QM), unaffected by $T$ or $P$.
  \item \textbf{Monotonicity:} Increasing severity or exposure, or degrading $T$ or $P$ (raising $C_u$), never decreases ASIL.
  \item \textbf{Bounded composition within $C_u$:} Strong $T$ can partially absorb weakened $P$ inside the Controllability-evidence report, but $C_u$ is never used to relax ASIL elsewhere in the safety case (see the double-counting prohibition of Section~\ref{sec:Cu_computation}).
\end{itemize}

\subsection{Implementation}

During Hazard Analysis and Risk Assessment (HARA), practitioners record evidence for each sub-dimension:
\begin{itemize}
  \item \textbf{Transferability:} the evidence tuple $E_T$ of equation~\eqref{eq:E_T} including validated FTTI, redundancy data, fault-injection results, lower confidence bounds $\hat{p}_{\mathrm{succ}}^{\mathrm{LCB}}$, and the gap $\Delta T$ of equation~\eqref{eq:DeltaT}.
  \item \textbf{Predictability:} the vector monitor $\mathcal{M}_P$ of equation~\eqref{eq:vectorMP} with the calibration declarations of Section~\ref{sec:pred_calibration}, conformity metrics, observer-study data, or telematics comparison results.
\end{itemize}
The computed $C_u$ is reported in the Controllability column of the ISO~26262-3:2018 Clause~6 Table~4 ASIL determination as supplementary evidence informing the $C$ classification. Concretely, for a hazardous event characterized by Severity~$S$ and Exposure~$E$, the safety case records $(S, E, C_u)$ together with the underlying $E_T$, $\mathcal{M}_P$, $\Delta T$, and the calibration declarations of Sections~\ref{sec:Transferability_classes} and~\ref{sec:pred_calibration}, so that any reviewer can trace the Controllability classification back to the underlying fault-injection and observer evidence rather than to an opaque $C$-class assignment.

\subsection{Outcome}

This extended ASIL determination method embeds AV-specific attributes within the established ISO~26262 risk structure. It quantifies both the vehicle's intrinsic capability to manage hazards and its clarity of interaction with the environment, providing auditable, evidence-based inputs for functional safety assessment without altering the underlying ASIL methodology.

\section{Proposed Amendments to ISO 26262}
\label{sec:amendments}
\textbf{Note:} The proposed amendments are the authors' individual research recommendations to the ISO standards body and do not reflect any organization's compliance obligations.

This section outlines concise modifications across ISO~26262 Parts~1--10 proposed to operationalize the enhanced Controllability framework.  The goal is not to replace existing processes but to introduce measurable evidence paths for \textit{Transferability} and \textit{Predictability}, ensuring that functional safety analyses remain applicable to SAE~L3--L5 autonomous vehicles and explicitly include external road-user safety. The amendments are not intended to alter the standard's normative HARA, ASIL, or item-definition semantics, but rather to extend its applicability and evidence base.  

\subsection{General Amendment Principles}

Amendments follow four principles:
(1) extend the scope of safety to include external road users;
(2) integrate Transferability and Predictability where Controllability is referenced, supported by the per-demand evidence tuple $E_T$ of \eqref{eq:E_T} and the gap $\Delta T$ of \eqref{eq:DeltaT};
(3) reinforce architectural requirements for autonomous fallback and fault-tolerant design; and
(4) enable data-driven verification using simulation, naturalistic driving datasets, and human-observer studies.
These principles ensure continuity with ISO~26262's structure while expanding its evidence base.

\subsection{Part 1 -- Vocabulary}

Add formal definitions for \textbf{Transferability}, \textbf{Predictability}, \textbf{Fallback Mechanism}, \textbf{External Road User}, and the gap $\Delta T$.
Clarify that a \textit{Safe State} must balance occupant and external-user safety and that the \textit{Fault-Tolerant Time Interval (FTTI)} also bounds the period within which the AV must complete reasonably safe fallback to protect nearby actors.

\subsection{Part 2 -- Functional Safety Management}

Recommend dedicated organizational roles for external-road-user safety and cross-team interfaces linking functional-safety, human-factors, and autonomy disciplines.
Impact analyses for design changes shall evaluate effects on $E_T$ and on the four channels of the Predictability monitor $\mathcal{M}_P$.

\subsection{Part 3 -- Concept Phase}

Update Hazard Analysis and Risk Assessment (HARA) to record both Transferability ($T$) and Predictability ($P$) ratings as a part of Controllability ($C$), alongside Severity ($S$) and Exposure ($E$). For each hazard, the work product carries $E_T$, $\mathcal{M}_P$, the derived $P$-class, $C_u$, and the calibration declarations of Sections~\ref{sec:Transferability_classes} and~\ref{sec:pred_calibration}.
Safety goals can be a function of Predictability and fallback-performance requirements, including the response margin $\Delta t_{\mathrm{response}}$-validated through empirical testing and observer studies.

\subsection{Part 4-6 -- Product Development}

\textbf{System level:} decompose safety goals into verifiable requirements addressing fallback initiation timing, FTTI compliance, observer-facing intent clarity, and signal consistency.
\textbf{Hardware level:} define redundancy and diversity evidence criteria consistent with the claimed Transferability class, reported with their lower confidence bounds.
\textbf{Software level:} propose isolation between primary and fallback functions, and motion-planning constraints whose effect on the four Predictability channels is independently monitored (Section~\ref{sec:planner}).

\subsection{Part 7-8 -- Production and Supporting Processes}

Operational data collection includes monitoring outputs for Transferability and Predictability under field conditions, with explanations for any channel excursions. Tool qualification and change-management clauses reference the datasets, simulators, and observer models used for $\mathcal{M}_P$ and $\hat{p}_{\mathrm{succ}}$ calibration, thereby making the calibration basis itself traceable.

\subsection{Part 9 -- Safety Analyses}

Controllability in risk classification becomes an \textit{enhanced parameter} represented by $C_u$, derived from $T$ and $P$ via \eqref{eq:Cu}--\eqref{eq:DeltaP} and reported alongside the standard Controllability entry. Dependent-failure analysis must consider faults that simultaneously degrade fallback reliability and behavioral anticipatability, since a single root cause may compromise both.

\subsection{Part 10 -- Informative Guidance}

Introduce a new appendix on \textbf{Predictability Operationalization}, providing:
(i) the calibration protocols of Section~\ref{sec:pred_calibration};
(ii) recommended scenario families-unprotected turns, crosswalk yields, highway merging, work-zone narrowing-following the scenario-based evaluation discipline of ISO~34502:2022~\cite{ISO34502-2022} and the ADS V\&V guidance of ISO/TR~4804:2020~\cite{ISO4804-2020}; and
(iii) reporting templates that record the evidence tuple $E_T$ of \eqref{eq:E_T}, the vector monitor $\mathcal{M}_P$ of \eqref{eq:vectorMP}, the gap $\Delta T$ of \eqref{eq:DeltaT}, and the calibration basis for each declared $(o, s, f, a)$. This appendix provides practical guidance on implementing human-perception-based safety validation in ISO~26262 projects.

\subsection{Summary}

These concise amendments integrate Transferability and Predictability throughout the ISO~26262 lifecycle, extending the standard's applicability from driver-controlled to autonomous systems. The revisions preserve ISO~26262's core philosophy while embedding measurable, evidence-based assurance of both internal fallback performance and external behavioral intelligibility.

\section{Planner Design Implications}
\label{sec:planner}
The four channels of the Predictability monitor $\mathcal{M}_{P}$ introduced in Section~\ref{sec:pred_channels}, together with the Transferability indicators of Section~\ref{sec:Transferability}, can inform planner design. We note, however, that they should not be confused with safety evidence. A planner may employ contextual conformity, intent clarity, signal consistency, or kinematic-surprise penalties as candidate-trajectory filters, as terms in a constrained-optimization objective, or as reward-shaping signals for a learned policy. Such use can encourage smoother, more legible, and better-signaled behavior during development, and the monitor declarations of Section~\ref{sec:pred_reference}-reference model, observer class, ODD slice, calibration criterion, transfer directly to the planner's specification of its own objective. Similarly, the Transferability indicators-initiation latency $t_{\mathrm{init}}$, MRM success probability $\hat{p}_{\mathrm{succ}}$, and stop-zone compliance $z_{\mathrm{stop}}$-can inform fallback trigger thresholds and redundancy design; for instance, a planner aware of $\tau_{\mathrm{FTTI}}$ can maintain a continuously feasible fallback trajectory to reduce $t_{\mathrm{init}}$ at demand time.

Nevertheless, planner-internal rewards, costs, and constraints are not independent evidence that the executed behavior is predictable or that fallback capability is achieved. Three failure modes are of particular concern. First, a learned policy optimized against $S_{\mathrm{intent}}$ may commit early to one goal interpretation in ways that are locally legible but difficult to abort-surrogate overfitting that improves a scalar metric while leaving real observer-facing ambiguity unresolved. 

Second, a reward derived from $Q_{\mathrm{conf}}$ is only meaningful within the declared ODD slice; outside it, the planner may receive reward signals that are uncorrelated with actual anticipatability. Third, and most critically, using $\mathcal{M}_{P}$ as both a training signal and a safety-case evidence artifact introduces a circular dependency that invalidates the independence assumption underlying the safety argument. The evidence tuple $E_T$ of~\eqref{eq:E_T} should be populated from independent fault-injection campaigns and hardware-in-the-loop evaluation, not from simulations that share assumptions with the primary driving function; a $\Delta_T$ computation drawn from the same environment used to train fallback-awareness constraints is not a valid $\Delta_T$ measurement.

Any planner who uses Transferability or Predictability during optimization must therefore be evaluated by an independent monitor on executed closed-loop behavior, with the safety case reporting the monitor outputs, calibration basis, detection power, false-positive rate, and ODD-sliced failure diagnosis separately from the planner's objective. We emphasize that this independence is not a procedural nicety: it is the only mechanism that distinguishes a predictable planner from one that has merely learned to satisfy its surrogate.

\section{Conclusion}
\label{sec:conclusion}
This paper enhances the ISO 26262 functional safety framework by extending its Controllability construct to address the operational realities of autonomous vehicles. Rather than replacing established principles, the proposed approach decomposes Controllability into two measurable and complementary sub-dimensions: \textbf{Transferability}, representing the autonomous system’s intrinsic fallback capability, and \textbf{Predictability}, representing the degree to which external road users can anticipate its behavior. Together, these extensions preserve the intent of ISO~26262:2018 while enabling its direct application to SAE~L4-L5 automation.

The framework developed in this document aligns with ISO 26262’s risk factors: Severity, Exposure, and Controllability. However, it enhances Controllability by using measurable indicators grounded in system reliability, human factors science, and behavioral data. Mathematical definitions and validation methods are used to connect Transferability to fault-tolerant fallback performance and Predictability to trajectory clarity and human understanding. This is compatible with the ISO 26262:2018 standard and broadens risk assessment to include both the system's internal control and human understanding.

Comprehensive lifecycle integration points are identified throughout ISO~26262:2018 Parts 1-10, ensuring that evidence of fallback performance and behavioral intelligibility is incorporated into standard work products. These targeted amendments harmonize functional safety with ISO 21448:2022 SOTIF and bridge classical safety engineering, human-robot interaction research, and real-world autonomous vehicle validation.

By embedding Transferability and Predictability within the ISO~26262 structure, this work provides a scalable, evidence-driven pathway for certifying the safety of highly automated and fully autonomous systems. The proposed enhancements maintain the rigor of functional safety while aligning its assurance framework with the distributed, data-intensive, and human-interactive nature of modern autonomous driving.

\textbf{Limitations.} The numerical targets of Tables~\ref{tab:T_acceptance} and~\ref{tab:Predictability_classes}, and the per-channel thresholds of \eqref{eq:max_agg}, are scenario-family-, ODD-slice-, and observer-class-specific; they must be reported with their calibration basis and are not transferable across declarations without re-calibration. Naturalistic driving data may encode behaviors that may not reflect optimal safety practices and culturally specific norms-it is an evidence model, not a normative target. Where human-subject evidence is unavailable, an observer model serves as a reasonably approximate model, subject to validation and project-specific calibration, for $p(g\mid\xi_{0:t},c,a)$, with the substitution and residual uncertainty declared as part of the safety case. The framework does not propose ASIL relaxation based on high $T$ or $P$; the unified Controllability evidence class $C_u$ supplements, rather than replaces, the normative ISO~26262 Controllability parameter. High-confidence Transferability claims for safety-critical fault families require statistical care: a finite fault-injection campaign cannot establish lower confidence bounds beyond a level dictated by sample size, and rare-event estimation, accelerated testing, and scenario-coverage arguments are required for the higher classes of Tables~\ref{tab:T-classes} and~\ref{tab:T_acceptance}. Both $\mathcal{M}_P$ and $T_{\mathrm{achievable}}$ may degrade silently outside their declared ODD slices; the out-of-distribution rejection rule required by Section~\ref{sec:pred_reference} is part of the calibration declaration, not an optional extension.

\textbf{Future Work.} In the future, we are interested in the calibration methodology for Sections~III and IV - ODD-slice-specific thresholds,  observer-model validation against human-subject data,  and rare-event estimation for high-confidence class assignments remain open for the field. The composition policies of Table~VI merit empirical comparison against hazardous-event data to replace the illustrative $C_u$ convention adopted here.  Finally, the coupling between legibility-oriented planning and the $\Delta T$ gap warrants explicit study, provided that the independence requirement of Section~VIII is preserved throughout.

\textbf{DISCLAIMER}
This publication reflects the individual research views of the named authors and does not constitute the official position, policy, or design philosophy of any author's employer or affiliated organization. The proposed frameworks, mathematical formulations, acceptance criteria, and amendment recommendations are research contributions offered for consideration by standards bodies and the broader research community; they do not describe the current or intended internal processes, safety cases, compliance posture, or product architecture of any author's employer.

For the avoidance of doubt, this publication is not intended and shall not be construed to establish a maximum or minimum requirement for dependability, safety, or performance for any automated driving system, nor shall it restrict any organization from adopting approaches that differ from those described herein. Its contents may require revision as technology, regulatory guidance, and community understanding evolve.

Where the views of any individual author may appear to conflict with the positions of their employer, the employer's own published materials, official documentation, and design decisions shall prevail.

\bibliographystyle{IEEEtran}
\bibliography{References.bib}

@standard{ISO26262-1-2018,
  title        = {Road vehicles — Functional safety (ISO 26262)},
  organization = {International Organization for Standardization},
  year         = {2018},
  url          = {https://www.iso.org/standard/43464.html},
  note         = {ISO 26262 series}
}

@standard{ISO21448-2022,
  title        = {Road vehicles — Safety of the intended functionality (SOTIF) (ISO 21448:2022)},
  organization = {International Organization for Standardization},
  year         = {2022},
  url          = {https://www.iso.org/standard/77490.html}
}

@inproceedings{Dragan2013LegibilityHRI,
  author       = {Anca D. Dragan and Kenton C. T. Lee and Siddhartha S. Srinivasa},
  title        = {Legibility and Predictability of Robot Motion},
  booktitle    = {ACM/IEEE Int. Conf. on Human-Robot Interaction (HRI)},
  year         = {2013},
  doi          = {10.1109/hri.2013.6483603},
}

@phdthesis{Dragan2015Legible,
  author     = {Dragan, Anca},
  title      = {Legible Motion for Robot Planning and Control},
  school     = {Carnegie Mellon University},
  year       = {2015}
}

@misc{NHTSA-NDS-Data,
  author       = {{National Highway Traffic Safety Administration (NHTSA)}},
  title        = {Strategic Highway Research Program 2 (SHRP 2) Naturalistic Driving Study (NDS) Data},
  year         = {various years},
  howpublished = {Dataverse},
  publisher    = {Virginia Tech Transportation Institute},
  doi          = {10.15787/VTT1/HRCECD}
}

@standard{SAEJ2944-2015,
  author       = {{Society of Automotive Engineers (SAE) International}},
  title        = {Road Vehicle -- Human-Centric Driving Data Acquisition for Research and Development},
  year         = {2015},
  note         = {SAE J2944\_201503. Standard issued 2015-03-31.}
}

@techreport{NHTSA2021_ADSCommunication,
  author      = {James Jenness and Amy K. Benedick and Jeremiah P. Singer and Sarah Yahoodik and Elizabeth Petraglia and Joshua Jaffe and John M. Sullivan},
  title       = {Automated Driving Systems’ Communication of Intent with Shared Road Users},
  institution = {U.S. Department of Transportation, National Highway Traffic Safety Administration},
  number      = {DOT HS 813 148},
  year        = {2021},
  month       = {11},
  doi         = {10.21949/1530235}
}

@article{Schmitt2022_CanCarsGesture,
author={Schmitt, Paul and Britten, Nicholas and Jeong, JiHyun and Coffey, Amelia and Clark, Kevin and Kothawade, Shweta Sunil and Grigore, Elena Corina and Khaw, Adam and Konopka, Christopher and Pham, Linh and Ryan, Kim and Schmitt, Christopher and Frazzoli, Emilio},
  journal={IEEE Robotics and Automation Letters}, 
  title={Can Cars Gesture? A Case for Expressive Behavior Within Autonomous Vehicle and Pedestrian Interactions}, 
  year={2022},
  volume={7},
  number={2},
  pages={1416-1423}
}

@inproceedings{dragan2016planning,
  title        = {Planning for Autonomous Cars that Leverage the Effects on Human Drivers},
  author       = {Sadigh, Dorsa and Sastry, S. Shankar and Seshia, Sanjit A. and Dragan, Anca D.},
  booktitle    = {Proceedings of Robotics: Science and Systems (RSS)},
  year         = {2016},
  doi          = {10.15607/RSS.2016.XII.029}
}

@inproceedings{dragan2015effects,
  title        = {Effects of Robot Motion on Human-Robot Collaboration},
  author       = {Dragan, Anca D. and Bauman, Simon and Forlizzi, Jodi and Srinivasa, Siddhartha S.},
  booktitle    = {Proceedings of the 10th ACM/IEEE International Conference on Human-Robot Interaction (HRI)},
  year         = {2015},
  pages        = {51--58},
  doi          = {10.1145/2696454.2696473}
}

@inproceedings{schmitt2023roadahead,
  author={Block, Avram and Joshi, Swapna and Tabone, Wilbert and Pandya, Aryaman and Lee, Seonghee and Patil, Vaidehi and Britten, Nicholas and Schmitt, Paul},
  booktitle={2023 32nd IEEE International Conference on Robot and Human Interactive Communication (RO-MAN)}, 
  title={The Road Ahead: Advancing Interactions between Autonomous Vehicles, Pedestrians, and Other Road Users}, 
  year={2023},
  pages={16-23},
  doi={10.1109/RO-MAN57019.2023.10309535}
}

@article{dragan2014legibilityToo,
  author    = {Anca D. Dragan and Siddhartha S. Srinivasa},
  title     = {Integrating Human Observer Inferences into Robot Motion Planning},
  journal   = {Autonomous Robots},
  year      = {2014},
  volume    = {37},
  number    = {4},
  pages     = {351--368},
  doi       = {10.1007/s10514-014-9408-0},
}

@inproceedings{dragannikolaidis2016viewpoint,
  author    = {Stefanos Nikolaidis and Anca D. Dragan and Siddhartha S. Srinivasa},
  title     = {Viewpoint-Based Legibility Optimization},
  booktitle = {Proceedings of the 11th ACM/IEEE International Conference on Human-Robot Interaction (HRI)},
  year      = {2016},
  pages     = {271--278},
  doi       = {10.1109/HRI.2016.7451761},
}

@inproceedings{draganzhou2017expressive,
  author    = {Alice Zhou and Dylan Hadfield-Menell and Abhishek Nagabandi and Anca D. Dragan},
  title     = {Expressive Robot Motion Timing},
  booktitle = {Proceedings of the 2017 ACM/IEEE International Conference on Human-Robot Interaction (HRI)},
  year      = {2017},
  pages     = {22--31},
  doi       = {10.1145/2909824.3020200},
}

@standard{UL4600,
  author       = {{Underwriters Laboratories (UL)}},
  title        = {{UL~4600: Standard for Safety for the Evaluation of Autonomous Products}},
  year         = {2023},
  organization = {Underwriters Laboratories},
  address      = {Northbrook, IL, USA},
  note         = {Covers safety case development, behavioral transparency, and system-level assurance for autonomous systems}
}

@article{garikapati2024comprehensive,
  title={A comprehensive review of parallel autonomy systems within vehicles: applications, architectures, safety considerations and standards},
  author={Garikapati, Divya and Poovalingam, Sundaresan and Hau, William and De Castro, Ricardo and Shinde, Chaitanya},
  journal={IEEE Access},
  year={2024},
  publisher={IEEE}
}

@standard{IEEE2846-2022,
  author       = {{IEEE}},
  title        = {{IEEE Standard for Assumptions in Safety-Related Models for Automated Driving Systems}},
  organization = {Institute of Electrical and Electronics Engineers},
  year         = {2022},
  number       = {IEEE Std 2846-2022},
  doi          = {10.1109/IEEESTD.2022.9761121},
  url          = {https://standards.ieee.org/ieee/2846/10831/}
}

@article{SalayCzarnecki2018ML,
  author       = {Rick Salay and Krzysztof Czarnecki},
  title        = {Using Machine Learning Safely in Automotive Software: An Assessment and Adaption of Software Process Requirements in {ISO} 26262},
  journal      = {arXiv preprint arXiv:1808.01614},
  year         = {2018},
  url          = {https://arxiv.org/abs/1808.01614}
}

@inproceedings{Schroeder2011PedestrianGap,
  author       = {Bastian J. Schroeder and Nagui M. Rouphail},
  title        = {Estimating Pedestrian Behavior at Crosswalks: Stated Preference and Behavioral Models for Engineering Applications},
  booktitle    = {Transportation Research Record},
  year         = {2011},
  volume       = {2264},
  number       = {1},
  pages        = {90--98},
  doi          = {10.3141/2264-11}
}

@book{HCM2010Pedestrian,
  author       = {{Transportation Research Board}},
  title        = {Highway Capacity Manual},
  year         = {2010},
  edition      = {5th},
  publisher    = {National Academies Press},
  address      = {Washington, DC},
  note         = {Pedestrian perception--response time guidance, Chapter~17.}
}

@standard{ISO34502-2022,
  title        = {Road vehicles --- Test scenarios for automated driving systems --- Scenario based safety evaluation framework},
  organization = {International Organization for Standardization},
  year         = {2022},
  number       = {ISO 34502:2022},
  url          = {https://www.iso.org/standard/78951.html}
}

@techreport{ISO4804-2020,
  author       = {{International Organization for Standardization}},
  title        = {Road vehicles --- Safety and cybersecurity for automated driving systems --- Design, verification and validation},
  institution  = {International Organization for Standardization},
  year         = {2020},
  number       = {ISO/TR 4804:2020},
  url          = {https://www.iso.org/standard/80363.html}
}

\end{document}